\pdfoutput=1

\documentclass[11pt]{article}

\usepackage[table]{xcolor}

\usepackage[preprint]{acl}

\usepackage{times}
\usepackage{latexsym}

\usepackage[T1]{fontenc}

\usepackage[utf8]{inputenc}

\usepackage{microtype}

\usepackage{inconsolata}

\usepackage{booktabs}
\usepackage{xspace}
\usepackage{microtype}
\usepackage{amsfonts}
\usepackage{amsmath}
\usepackage{amssymb}
\usepackage{amsthm}
\usepackage{booktabs}
\usepackage{dsfont}
\usepackage{paralist}
\usepackage{times}
\usepackage{xspace}
\usepackage{multirow}
\usepackage{graphicx}
\usepackage{float} 
\usepackage{lipsum}
\usepackage{appendix}
\usepackage{enumitem}
\usepackage{pifont}
\usepackage{makecell}
\usepackage{caption}
\usepackage{latexsym}
\usepackage{cleveref}
\usepackage{subcaption}
\usepackage{placeins}
\usepackage{upgreek}

\usepackage[export]{adjustbox}

\crefformat{section}{\S#2#1#3}
\crefformat{figure}{Fig.~#2#1#3}
\crefformat{table}{Tab.~#2#1#3}
\crefformat{appendix}{\S#2#1#3}

\newcommand{\ra}[1]{\renewcommand{\arraystretch}{#1}}
\definecolor{lightgray}{gray}{0.95}
\usepackage{soul}

\newcommand\dataset{\textsc{Reveal}\xspace}

\renewcommand\paragraph[1]{
\vspace{0.15cm}
\noindent 
\textbf{#1}
}

\title{A Chain-of-Thought Is as Strong as Its Weakest Link: \\  A Benchmark for Verifiers of Reasoning Chains}

 \author{Alon Jacovi$^{1,2}$ \ Yonatan Bitton$^1$ \ Bernd Bohnet$^3$ \ Jonathan Herzig$^1$ \\ [5px] \textbf{Or Honovich$^{1,4}$ \ Michael Tseng$^3$ \ Michael Collins$^3$ \ Roee Aharoni$^1$ \ Mor Geva$^{1,4}$} \\ [10px]
        $^1$Google Research \quad  $^2$Bar Ilan University \quad $^3$Google DeepMind  \quad  $^4$Tel Aviv University \\
        \tt alonjacovi@google.com}

\begin{document}
\maketitle
\begin{abstract}

Prompting language models to provide step-by-step answers (e.g., ``Chain-of-Thought'') is the prominent approach for complex reasoning tasks, where more accurate reasoning chains typically improve downstream task performance.
Recent literature discusses automatic methods to verify reasoning steps to evaluate and improve their correctness. However, no fine-grained step-level datasets are available to enable thorough evaluation of such verification methods, hindering progress in this direction.
We introduce \dataset: \textit{Reasoning Verification Evaluation}, a new dataset to benchmark automatic verifiers of complex Chain-of-Thought reasoning in open-domain question answering settings.
\dataset includes comprehensive labels for the relevance, attribution to evidence passages, and logical correctness of each reasoning step in a language model's answer, across a wide variety of datasets and state-of-the-art language models. Available at \url{reveal-dataset.github.io}.

\end{abstract}

\section{Introduction}

\begin{figure}[t]
\setlength{\belowcaptionskip}{-15pt}
\centering
\includegraphics[width=0.99\linewidth]{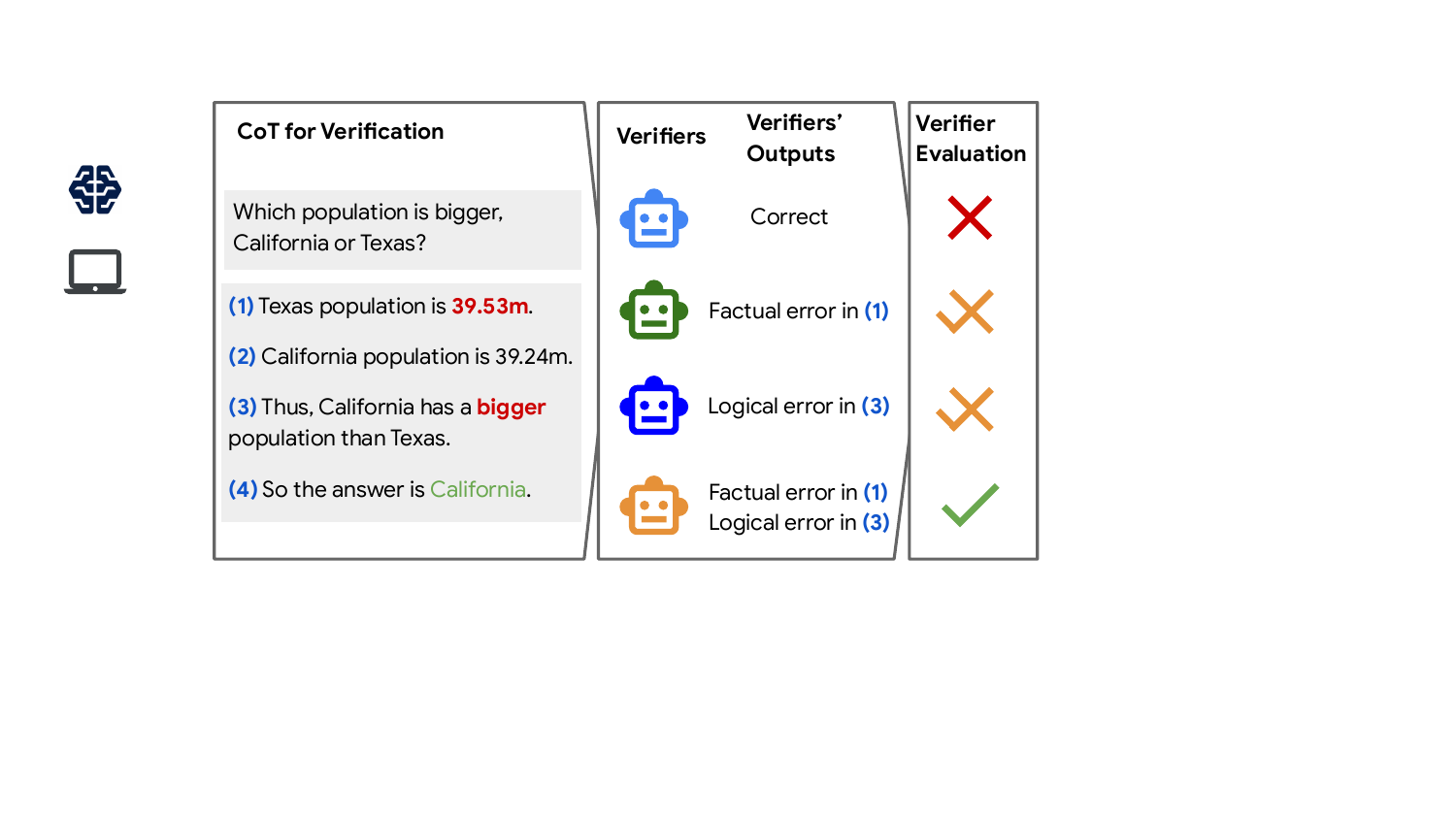}
\caption{We collect \dataset, an evaluation benchmark for the task of verifying reasoning chains in Chain-of-Thought format, which checks whether a reasoning chain is a correct justification to the final answer (importantly, the answer can be correct even if the reasoning is incorrect, as in the example above).  The figure shows four verifiers (middle) verifying the correctness of a CoT (left). We use the dataset to benchmark multiple verifiers (right).
}
\label{fig:teaser}
\end{figure}

Complex reasoning tasks involve answering questions that require multiple steps of reasoning \cite{welbl-etal-2018-constructing,talmor-berant-2018-web}. Addressing these questions may require open-domain knowledge \cite{geva2021strategyqa}, mathematical reasoning \cite{DBLP:journals/corr/abs-2110-14168gsm8k,Hendrycks2021MeasuringMP}, logic \cite{dalvi-etal-2021-explaining}, and so on.
Reasoning chains---breaking the task into multiple steps explicitly---is useful for improving performance in such tasks, with LMs demonstrating better performance when encouraged to generate the reasoning chain behind their answer \cite{DBLP:conf/icml/LampinenRDCTMYS22,zelikman2022star,hu2023visual}, commonly implemented via Chain-of-Thought (CoT) prompting \cite[see \Cref{fig:teaser} for an example]{wei2023chainofthought}. 

Evaluation in such settings is traditionally limited to evaluating only whether the final answer is correct \cite{chowdhery2022palm,wei2023chainofthought,DBLP:conf/iclr/0002WSLCNCZ23}. 
However, correct reasoning chains have been shown to be correlated with better final answers \cite{jung-etal-2022-maieutic}, with recent literature proposing automatic methods for verifying the quality of the reasoning chains themselves along various axes such as informativeness, relevance, factuality and logical correctness \cite{DBLP:conf/iclr/GolovnevaCPCZFC23,press2023measuring,opitz-frank-2021-towards,Leiter2022TowardsEE}. While such verification methods are a promising direction for improving reasoning in LLMs, it is not clear how to evaluate them due to the lack of high-quality, step-level annotated data, and collecting such data was shown to be difficult (in terms of reaching high inter-annotator agreement) and costly \cite{DBLP:conf/iclr/GolovnevaCPCZFC23}.

\begin{figure}[t]
\setlength{\belowcaptionskip}{-10pt}
\centering
\includegraphics[width=0.75\linewidth]{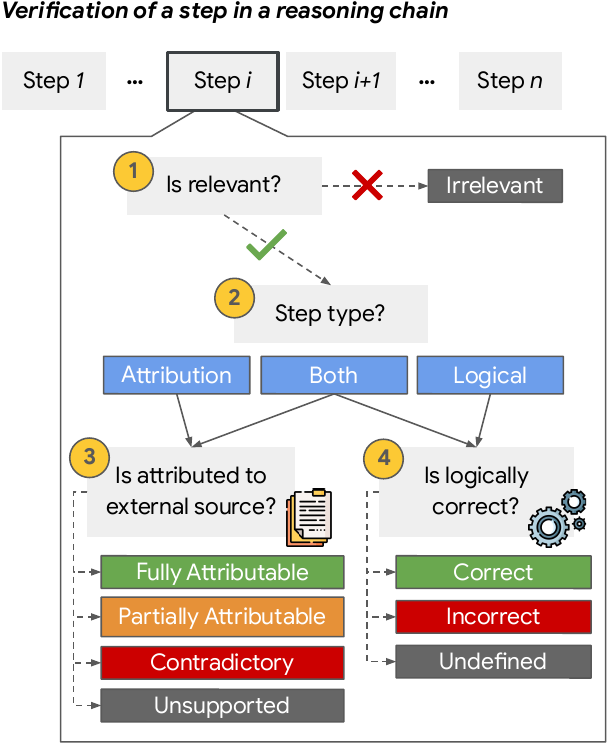}
\caption{A flowchart of our protocol for verifying reasoning correctness step-by-step (\Cref{sec:formalism}).}
\label{fig:flowchart}
\end{figure}

\begin{figure}[t]
\setlength{\belowcaptionskip}{-10pt}
\centering
\includegraphics[width=0.99\linewidth]{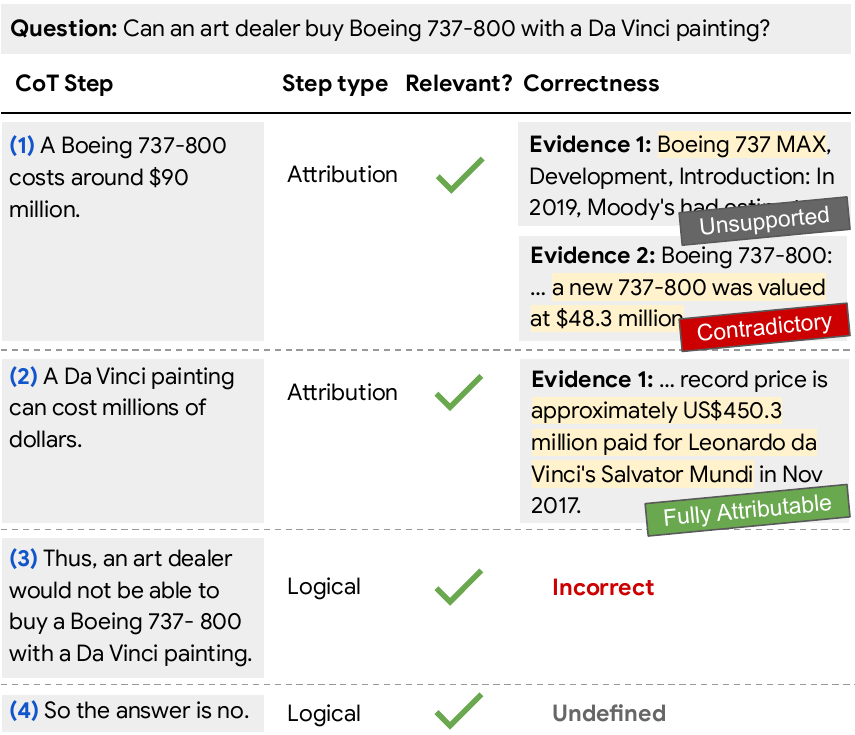}
\caption{A \dataset instance, with labels for step type, relevance, and correctness (attribution to a source or logical correctness from previous steps). 
Each label is accompanied by a high-quality free-text justification (shown in \Cref{fig:justifications}).
We retrieve up to three evidence paragraphs, until a non-``unsupported'' evidence is found. For Step (4) logical correctness is undefined, since it follows a logically incorrect step (Step 3).
}
\label{fig:data-example}
\end{figure}

We present \dataset (\textit{Reasoning Verification Evaluation}), an evaluation benchmark for complex reasoning verifiers.\footnote{At \url{huggingface.co/datasets/google/reveal}. We adopt practices by \citet{jacovi2023stop} against data contamination and request that any future redistribution or usage of the data respects the same constraints.} 
\dataset covers a diverse set of reasoning skills, complexity levels, and knowledge domains.
It contains 704 unique questions from 4 popular QA datasets, and 1,002 CoT answers generated by 3 language models, consisting of 3,360 CoT steps in total. 

\begin{figure}[t]
\setlength{\belowcaptionskip}{-10pt}
\centering
\includegraphics[width=0.99\linewidth]{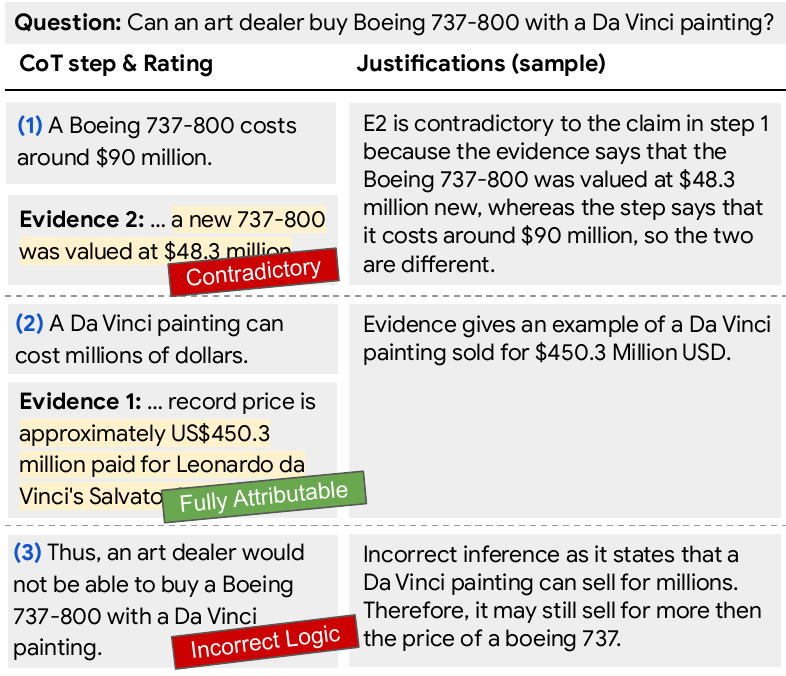}
\caption{Example free-text justifications for the labels in \dataset. There are 5 justifications for each label.
}
\label{fig:justifications}
\end{figure}

Each step is first labeled for relevance with respect to the final answer, and then whether the step is an attribution step (introduces factual knowledge which can be attributed to a source), a logical step (introduces logical inference from previous steps) or both. For attribution steps, we collect labels for correctness to retrieved Wikipedia paragraphs given as evidence (with full support, partial support, contradiction, or no-support as labels). For logical steps, we label for logical correctness. Each label includes free-text justifications written by the annotators. An illustrative instance from the dataset is shown in \Cref{fig:data-example,fig:justifications}. We split the dataset into \dataset-Eval, the main evaluation benchmark containing high inter-annotator-agreement labels, and \dataset-Open, a smaller set of interesting borderline cases with open labels due to low inter-annotator agreement. In \Cref{sec:the-dataset} we describe the dataset, and report fine-grained analyses of non-attributable steps in \dataset-Eval (i.e., evidence that supports or contradicts them was not found) and of disagreement categories in \dataset-Open.

\dataset supports versatile evaluation settings, for example: (1) Attribution steps, along with their evidence, can serve as a high-quality Natural Language Inference \cite[NLI,][]{DBLP:conf/mlcw/DaganGM05,bowman-etal-2015-large} benchmark in a setting of fact-checking LM outputs \cite{gao-etal-2023-rarr,zhang2023language}; (2) CoT verifiers can be evaluated at the level of individual steps, or (3) at the level of full CoT answers; (4) Each label in the data contains five free-text justifications (one per annotator), which can accommodate research around the generation of explanations and justifications, or be used to understand nuance in borderline cases.

As we focus on the evaluation of step-level validation in complex reasoning, in \Cref{sec:baselines} we report the performance of multiple up-to-date verification baselines,
leveraging NLI classifiers, GPT-3 and PaLM 2,
showing much room for improvement in current state-of-the-art solutions. In particular, verifiers struggle at classifying whether a step conveys correct logical inference from previous steps. 

In summary, this work includes the following contributions: (I) A protocol for step-by-step verification of reasoning chains (\S\ref{sec:formalism}); (II) An annotation schema to reliably execute the protocol with human annotators (\S\ref{sec:schema}); (III) A new benchmark dataset for evaluating automatic reasoning chain verifiers (\S\ref{sec:data-collection}, \S\ref{sec:the-dataset}); (IV) Detailed analyses of challenges in retrieving evidence to knowledge claims in reasoning and documentation of disagreements in the data (\S\ref{sec:the-dataset}); (V) A study on the challenges for current verifiers (\Cref{sec:baselines}). These contributions advance the research on verification of reasoning chains and methods for correctly reasoning about complex questions.

\section{Formalism for Verification of Reasoning Chains}

In this section, we formalize the task of verifying reasoning chains for attribution and logical correctness and discuss its evaluation.

\label{sec:formalism}

We consider a reasoning chain as a sequence of $n$ steps $r=s_1, ..., s_n$, where $s_i$ is a claim generated conditionally on the steps preceding it $s_1, ..., s_{i-1}$. 
We focus on reasoning chains that are generated to answer a given question $q$ and have a CoT format, that is, where the last step $s_n$ includes the final answer to $q$ and every step is a standalone sentence.

Given vocabulary $\mathbb{Z}$, a verifier $\mathcal{V}: \mathbb{Z}^{|q|} \times \mathbb{Z}^{|r|} \rightarrow [0, 1]$ 
receives a question $q$ and a reasoning chain $r$ conditioned on $q$, and outputs a score for the correctness of $r$ as an answer to $q$. 
``Correctness'' can be defined in various factors, such as factuality, grammaticality, and coherence. We focus on two key factors---correctness with respect to world knowledge, i.e., that the answer is derived based on grounded facts, and logical correctness, where reasoning steps are inferred with logically correct inference.
We consider the world knowledge that supports the claims in $r$ as some evidence $e$, external to the reasoning chain.
For our purposes, we separate the retrieval of $e$ from the verification process. 
Therefore, we consider a verifier $\mathcal{V}: \mathbb{Z}^{|q|} \times \mathbb{Z}^{|r|} \times \mathbb{Z}^{|e|} \rightarrow [0, 1]$ that also receives as input evidence $e$ to ground the correctness of $r$.

\paragraph{Full chain vs. step-level verification.} At high-level, a verifier is a system which receives as input a question and a reasoning chain answer (in our case, a CoT answer), and outputs whether the \textit{entire} reasoning chain is correct---this is the more prevalent approach (\Cref{subsec:related-work}). 
A more fine-grained approach is to evaluate each reasoning step for correctness separately, where the reasoning chain is correct if every step is correct \cite{li-etal-2023-making}. We adopt the step-level approach, as it provides the ability to evaluate full-chain verifiers, the ability to detect the exact point of failure in reasoning chains, and the ability to distinguish between different types of errors. Step-level detection can additionally help with detecting cases of ``snowballing of hallucinations'' from earlier steps \cite{zhang2023language}.

\paragraph{Correctness of an individual step.} \Cref{fig:flowchart} details our methodology for defining the correctness of an individual step, which we expand on below.

\vspace{0.1cm}
\noindent 
(1) \textit{Step relevance.} Each step is ``\verb+relevant+'' or ``\verb+irrelevant+'' to answer the question. Irrelevant steps do not invalidate the chain's correctness.

\vspace{0.1cm}
\noindent 
(2) \textit{Step type.} The correctness of a step can be defined in three ways: Whether it is entailed from approved world knowledge (``\verb+attribution step+''), entailed from previous steps (``\verb+logical step+''), or both (``\verb+both+''). We define attribution steps in language adopted from fact-verification  \cite{DBLP:journals/corr/abs-1809-08193,guo-etal-2022-survey}: If the step \textit{``contains knowledge that should be verified against an external source.''}

\vspace{0.1cm}
\noindent 
(3) \textit{Step attribution to external source.} An attribution step is correct if it is ``\verb+fully attributable+'' to a given source, meaning ``strictly according to the given source, all of the information in the claim is correct.''  Steps are otherwise ``\verb+contradictory+'' or ``\verb+partially attributable+'' to the source, or ``\verb+unsupported+'' by the source. This categorization mimics the Attribution to Identified Sources formalization \cite[AIS,][]{DBLP:journals/corr/abs-2112-12870}.

\vspace{0.1cm}
\noindent 
(4) \textit{Step logical correctness.} Each logical step is ``\verb+correct+'' if it can be logically inferred from the previous steps, otherwise it is ``\verb+incorrect+''. The correctness of logical steps that follow incorrect logical steps is undefined. 

\vspace{0.11cm}
\noindent 
(3 \& 4) \textit{Hybrid steps.} When steps contain both world knowledge and logical inference, we assume that all external knowledge is fully attributable for the purpose of logical correctness, and that all logical inferences from previous steps are correct for the purpose of attribution correctness.

\section{Annotation Schema} \label{sec:schema}

We introduce a labeling schema for reasoning chain verification, according to the formalization in \Cref{sec:formalism}. The schema prioritizes annotation quality, ease of use and scalability. Such an implementation requires making some design choices, which we explain and justify below. The annotation interfaces and annotation questionnaire are available in \Cref{app:data-collection}.

We observe that the overall annotation process of a reasoning chain involves different types of information and requires various skills. For example, to determine if a reasoning step can be attributed to a given evidence paragraph, it is essential to read the paragraph in detail and judge its relation to the step without any additional context. In contrast, to validate if a reasoning step is logically correct, it is required to think critically whether some information is missing or incorrect in the overall reasoning process while leveraging the annotators' background knowledge. The two modes of thinking (logic and attribution) can interfere with each other, where focusing on one at a time is less cognitively-demanding.

To accommodate the complexity of the verification task, we therefore split it into two annotation tasks: (1) A full-chain task that considers the logical transitions between steps, while assuming that any specified facts are factual; (2) a step-wise task that checks if a given step is attributable to or contradicts some given external knowledge.

Note that \textit{the two tasks are designed to be independent}, such that different annotators can annotate logical correctness independently from attribution, and vice versa. This allows for parallelization of the annotation process, reduces cognitive load, and provides a more robust set of labels that is less dependent on specific annotators. For example, some annotators were assigned to only one of the tasks, according to feedback during pilot phases.

\paragraph{Justifications.}
In addition to the verification labels, in both tasks we ask annotators to provide a free-text justification for their choices. This information helps to monitor the annotation process, provide additional valuable insight on the reasoning behind ratings---as it is possible to make the same decisions in these tasks for different reasons---and additionally serves as a method of interacting with the annotators to provide and receive feedback.

\subsection{Task 1: Relevance, Type, and Logic}

Given a complex reasoning question and a CoT answer, the first task involves annotating the chain step-by-step, labeling the following information for each step: \textit{relevance}, \textit{step type} and \textit{logical correctness}, according to the classes described in \Cref{sec:formalism}.

The logical correctness of a step in the chain is evaluated with respect to the steps preceding it and the question. Importantly, this is done without considering the correctness of attribution steps that introduce external knowledge---i.e., all external knowledge is assumed to be correct in this task, as the focus is on the logical derivations. Consider for example the following reasoning chain:
\begin{quote}
    \textit{Question:} Which population is bigger, California or Texas?

    \textit{Answer:} Texas population is 39.53m.$^{(1)}$ California population is 39.24m.$^{(2)}$ Thus, Texas has a bigger population than California.$^{(3)}$ So the answer is Texas.$^{(4)}$
\end{quote}
When evaluating logical correctness, step 3 would be considered correct, even though the cited number for the population of Texas is incorrect.

\subsection{Task 2: Relevance and Attribution}
The second task considers the correctness of facts stated in the reasoning chain and therefore is applied only to steps labeled as attribution steps (in the first task). The task involves verifying every step against a set of evidence paragraphs, where
the annotator should indicate for a given evidence paragraph if the step is fully attributable (supported), partially attributable, contradicts or is unsupported. The annotation of a step ends when either a fully-supporting or contradicting evidence is found, or the maximum number of paragraphs is labeled. In this work, we limit this number to three, as evidence support dramatically decreases with additional evidence (\Cref{fig:stats} right) and to reduce cost.

Relevance is annotated in this task as well, to account for more specific phrasing under the lens of attribution: An attribution step is marked relevant here if it introduces information that is helpful to answer the question.
When writing justifications to their ratings, annotators were encouraged to quote specific parts of the CoT or evidence as needed.

\section{Data Collection Process}
\label{sec:data-collection}
In this section, we describe the full process for constructing \dataset, using our annotation schema. The process is divided into two phases: Collecting and generating data for annotation (steps 1--2) and the annotation process (step 3). 

\paragraph{Step 1: Reasoning Chain Generation.}
We use four open-domain complex-reasoning QA datasets to elicit model-generated reasoning chains:
\begin{enumerate}
[leftmargin=*,topsep=3pt,parsep=0pt]
    \item \textsc{StrategyQA} \cite{geva2021strategyqa}: Yes/no questions that require a diverse set of reasoning skills and applying implicit knowledge.
    \item \textsc{MuSiQue} \cite{trivedi2021musique}: Multi-hop reasoning questions with free-text entity answers, generated based on paragraphs from Wikipedia. For our dataset, we consider the subset of 2-hop questions, as we found qualitatively that the 3-hop and 4-hop questions are often unnatural and hard to understand. 
    \item \textsc{Sports Understanding} \cite{srivastava2023imitation}: Yes/no questions that require reasoning about knowledge on sports players, leagues, and sports maneuvers.
    \item \textsc{Fermi} \cite{kalyan-etal-2021-much}: Estimation questions with numerical answers that require both knowledge and reasoning to answer. The questions are designed to have no clear gold answers---they are exercises of common-sense reasoning and numerical estimation, e.g., ``How much water does a school use in a week?''
\end{enumerate}
Examples from each source are provided in \Cref{tab:source-data-examples} (\Cref{app:data-collection}).
This diverse set of tasks provides a variety of question types, knowledge and reasoning requirements, and answer formats (binary, numerical and free-text entities). The questions were sampled randomly from the evaluation sets of each dataset and evenly split across the four datasets.

We use three different LMs to generate CoT answers: Flan-PaLM-540B \cite{chung2022scaling}, GPT-3 \cite[text-davinci-003,][]{DBLP:journals/corr/abs-2005-14165gpt3} and Flan-UL2-20B \cite{tay2023ul2}. We prioritized a variety in the models, with two large-size high-performing models with different pretraining data, and one smaller-size, weaker model, in order to gather a variety of CoT answers.
CoT prompt demonstrations were written according to standard practices \cite{wei2023chainofthought}, with in-domain examples for each dataset taken from the dataset's training set, and designed to be simple and informative.
A subset of the questions from each dataset (one-fourth) were answered by all three models with three separate CoT answers, to serve as a base for analyses that require multiple answers per question, while the rest were answered by each model once in equal proportions (one-fourth each). Our goal in this methodology was to maximize the flexibility of the dataset and potential for analyses and evaluations, for as many LMs and verifiers as possible.  Specifically, we aimed to collect a sufficient variety of answers, \textit{both correct and incorrect}, from a practical and realistic distribution.

\paragraph{Step 2: Evidence Retrieval.}
For attribution verification, we use Wikipedia as an external knowledge source and retrieve three paragraphs for each attribution step. StrategyQA, MuSiQue, and Sports Understanding are well-supported by Wikipedia, while Fermi is explicitly designed to be difficult to support, giving a variety of cases.
To promote the retrieval of supporting paragraphs, we mix dense retrieval and lexical-based retrieval, fetching two paragraphs with GTR \cite{ni2021large} and one with BM25 \cite{bm25variant}. In addition, retrieval is done using decontextualized versions of the CoT steps, generated by the decontextualization model of \citet{choi-etal-2021-decontextualization} which replaces co-referenced pronouns with explicit entity mentions.

\paragraph{Step 3: Annotation.}
We use a pool of 13 English-speaking annotators, and collect 5 annotations per question and its corresponding CoT answer, for each of our two tasks (see \Cref{sec:schema}).
A small portion of the labels (approx. 5\%) were annotated by the authors of this work (3 annotations per sub-task), to fill any gaps in examples that were not fully annotated for technical reasons.

\begin{table}[t]\centering
\setlength\belowcaptionskip{-8px}
\setlength\tabcolsep{3pt}
\scriptsize
\rowcolors{1}{}{lightgray} \ra{1.2}
\resizebox{0.99\linewidth}{!}{
\begin{tabular}{>{\raggedright}p{2.9cm}rrr}\toprule
&\textbf{\dataset-Eval} &\textbf{\dataset-Open}\\\midrule
Questions &704 &205 \\
CoT answers &1002 &224 \\
CoT steps &3360 &847 \\
Avg. steps per CoT & 3.4 & 5.1 \\
Attribution steps &1979 &485 \\
Step-evidence pairs &3502 &745 \\
Avg. evidence length (words) & 103 & 103 \\
Logic steps &1250 &306 \\
Fully attributable step-evidence pairs & \multirow{2}{*}{864} & \multirow{2}{*}{--} \\
Logically correct steps & 1063 & -- \\
Fully correct CoT answers & 200 & -- \\ 
\bottomrule
\end{tabular}}
\caption{Quantities for various properties of \dataset.}\label{tab:data-quantities}
\end{table}

\begin{figure*}[t]
\setlength{\belowcaptionskip}{-10pt}
\centering
\resizebox{0.99\linewidth}{!}{
\includegraphics[valign=t]{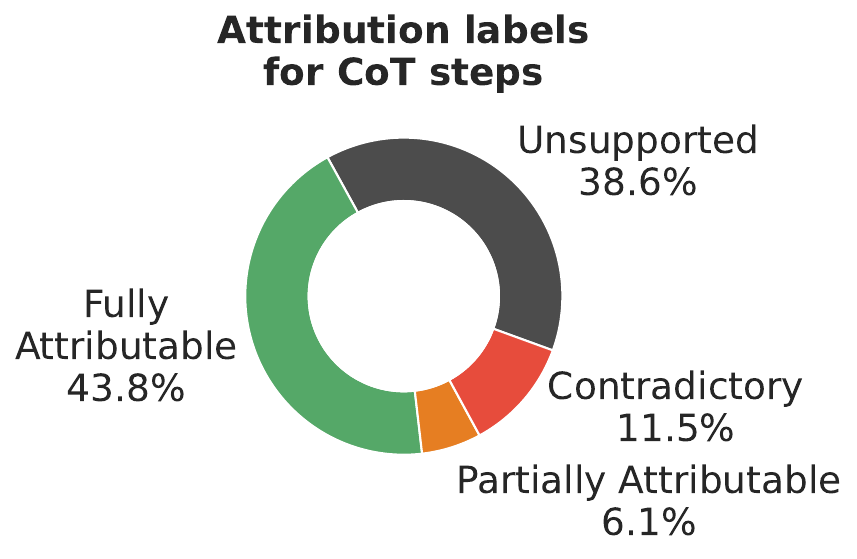}
\includegraphics[valign=t]{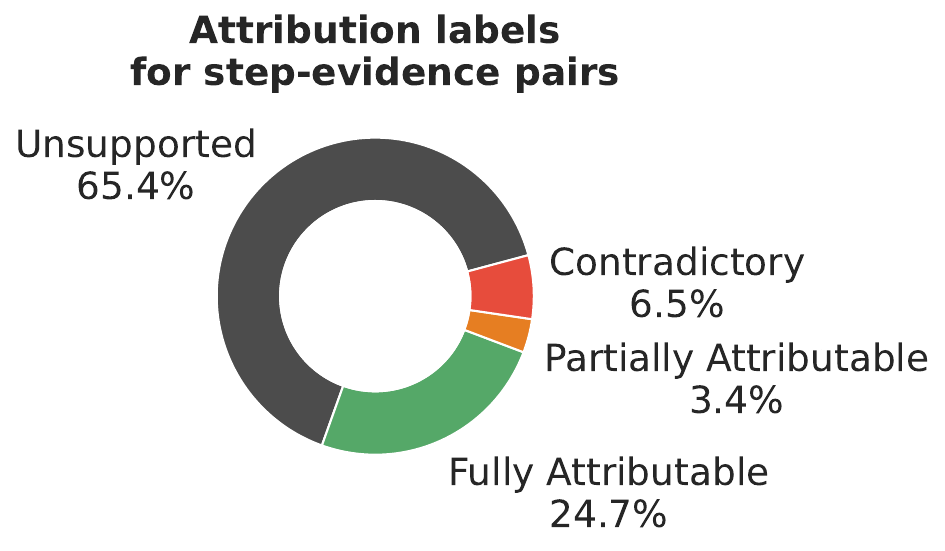}
\includegraphics[width=0.025\linewidth,valign=t]{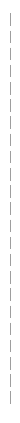}
\includegraphics[valign=t]{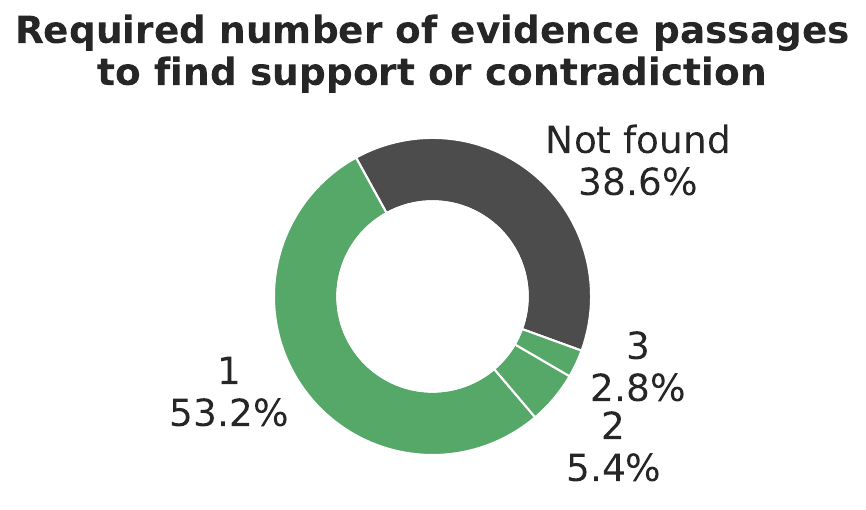}
}
\caption{Statistics for the attribution task in \dataset-Eval (\Cref{sec:the-dataset}). Left: Label distributions for each step or step-evidence pair. Right: Statistics for the number of retrieved evidence passages for each attribution step until a supporting or contradicting evidence was found (up to 3).
}
\label{fig:stats}
\end{figure*}

\paragraph{Justifications and Quality Validation.}
Each annotation in \dataset is accompanied by a free-text justification written by its annotator. Those justifications serve two goals: they allow us to easily monitor the annotation process by leaning into the annotators thought process and provide them with meaningful feedback \cite{nangia-etal-2021-ingredients}, and serve as a valuable resource on their own which we leave to be explored in future work. We maintained a high level of quality for these justifications, such that they do not collapse to templated answers or provide insufficient information. Examples of these justifications are given in \Cref{fig:justifications}.

In addition,
we performed three pilot rounds for annotator selection, annotator training, and task improvement, all to maintain a high quality bar for the collected dataset (the pilot annotations were discarded from the final dataset). Finally, we split the dataset into two subsets, \dataset-Eval and \dataset-Open, where the latter contains low-confidence labels that received lower inter-annotator agreement (see \Cref{sec:the-dataset}).

\section{\dataset} \label{sec:the-dataset}
We ran the data collection protocol described in the previous section to obtain our dataset, \dataset. In terms of inter-annotator agreement, we report a Krippendorf's $\alpha$ of 0.49 for attribution steps and 0.46 for logical steps.

We split \dataset into two subsets, as we observe that some reasoning chains are very challenging to annotate because of ambiguity or other factors (analyzed in  \Cref{subsec:dataset-hard-cases}).
Any CoT answer with at least one step that has low inter-annotator agreement is treated as an indecisive case, which applies to 18\% of the CoT answers. A step is considered to have ``low inter-annotator agreement'' if less than three annotators agree on any single label for attribution or logical correctness (for example, an attribution step labeled as ``partially attributable'' by two annotators, ``fully attributable'' by one annotator, and ``contradictory'' by two annotators). We release these examples separately in a data subset called  \textbf{\dataset-Open}, as they are valuable instances of difficult or borderline cases, and the rest of the examples in a subset called \textbf{\dataset-Eval}. Since the annotations in \dataset-Open are indecisive, in \Cref{sec:baselines} we evaluate only on \dataset-Eval.

\Cref{tab:data-quantities} details statistics on the two data splits, showing that \dataset-Eval has several hundreds to thousands of examples for each setting to support reliable evaluation.
Instances are approximately distributed uniformly across source datasets and LM generators. We release all five annotations for each label in the dataset, including anonymized annotator identifiers for each label, to support future annotation methodology research and reproducibility \cite{sandri-etal-2023-dont}.

\begin{figure}[t]
\setlength{\belowcaptionskip}{-10pt}
\centering
\includegraphics[width=0.45\linewidth,valign=t]{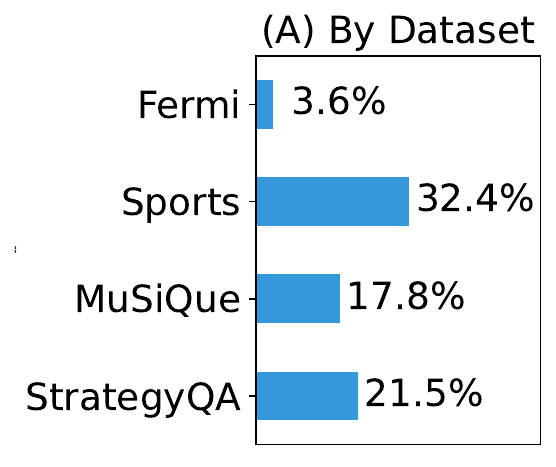}
\includegraphics[width=0.45\linewidth,valign=t]{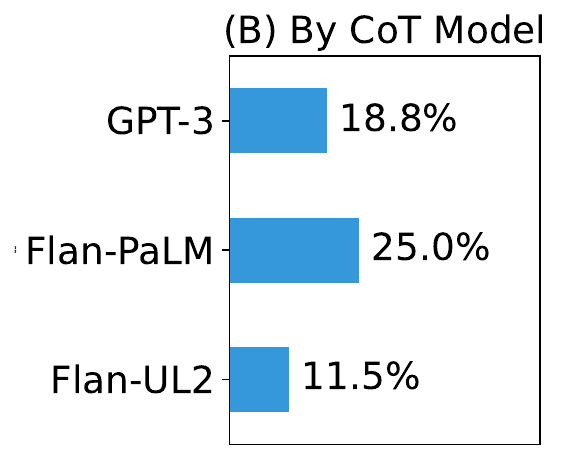}
\caption{Comparison of full CoT correctness across subsets in \dataset-Eval, meaning that \textit{all steps} in the CoT are fully attributable and logically correct. (B) only includes questions answered by all models, where all answer CoTs are in \dataset-Eval.
}
\label{fig:model_accuracy}
\end{figure}

Considering the label distributions, 98.6\% of steps are relevant to answer the question, and 87.5\% of logic steps are logically correct. 
\Cref{fig:stats} (left) shows label distributions for attribution steps and step--evidence pairs (note that partial attribution is considered as ``not enough info'' in standard NLI formalization). While 43.8\% of attribution steps were fully attributable, this is still relatively low, which may point at attribution as a main obstacle in reasoning. However, we do note that unsupported steps are not necessarily incorrect, as lack of support may stem from wrong claims, but also from issues in retrieving relevant supporting evidence. We analyze this further in \Cref{subsec:unsupported-analysis}.

\Cref{fig:stats} (right) shows the distribution for the required number of evidence paragraphs for finding support or contradiction for each step: In the majority of cases where support was found, it was already present in the first retrieved paragraph.

\Cref{fig:model_accuracy} shows distributions for full chain-level attribution and correctness for each subset of the data. In terms of specific types of errors in the CoTs: 77.3\% of CoTs in \dataset-Eval have a step which is not fully attributable, while 18.5\% have a step which is logically incorrect.

\subsection{Analysis of Unsupported Claims} \label{subsec:unsupported-analysis}
A large fraction of the labeled steps was found unsupported by the retrieved evidence (38.6\%, \Cref{fig:stats}). To understand this, we conduct a qualitative analysis of 40 unsupported steps (10 from each source dataset, randomly sampled). For each step, we first verify whether the step is factually correct. For correct steps, we analyzed whether the issue was due to irrelevant retrieved evidence, or due to other reasons. See \Cref{app:data-collection} for full details.

Of the 40 unsupported steps, 19 were indeed not factual. Of the remaining 21: In 13 steps the retrieved evidence was relevant, but additional reasoning and world knowledge was required to verify them (e.g., which teams are part of the spanish ``La-Liga'', or that ``ACM'' stands for ``Association for Computing Machinery''). In 6 steps the retrieved evidence was irrelevant. In 1 step there were two claims within the given step, but only one was supported by the evidence, and in 1 step the evidence was irrelevant due to insufficient decontextualization (``Brown'' ref. ``Jaylen Brown'').

From this analysis, we estimate that roughly half of the unsupported cases stem from using imperfect retrieval. Additional challenges such as imperfect decontextualization and requiring to reason over multiple evidence passages can also slightly contribute to failing to support factually correct claims.

\subsection{Analysis of \dataset-Open} \label{subsec:dataset-hard-cases}

\dataset-Open consists of questions paired with answers in which at least one of the steps has low inter-annotator agreement, defined as a case where no more than two (out of five) annotators agree on a single label for that step.
\Cref{tab:indecisive-cases} summarizes our analysis of the low-agreement cases. We inspected each of these steps (including the underlying data, as well as annotators' individual labels and written feedback) and report a qualitative description of the complications that may have contributed to the low agreement. On the basis of these qualitative notes, we devise 13 broader categories and manually assign one or more categories to each step.

\begin{table}[t]\centering
\setlength\belowcaptionskip{-8px}
\setlength\tabcolsep{3pt}
\scriptsize
\rowcolors{1}{}{lightgray} \ra{1.2}
\resizebox{0.99\linewidth}{!}{
\begin{tabular}{
>{\raggedright}p{3.5cm}
>{\raggedleft\arraybackslash}p{1.8cm}
>{\raggedleft\arraybackslash}p{1.2cm}
}
\toprule
\multirow{1.8}{*}{\textbf{Complexity Category}} & \textbf{Attribution steps} (N=199) & \textbf{Logic steps} (N=52) \\\midrule
Approximating and hedging & 1.51\% & -- \\
Averages and ranges & 10.05\% & -- \\
Calculation or unit issue & -- & 42.31\% \\
Specialized knowledge & 13.07\% & 7.69\% \\
Formatting issue & 3.02\% & -- \\
Unclear reference & 12.56\% & 30.77\% \\
Inference across claims in evidence & 1.51\% & -- \\
Invalid inference in previous step & 0.50\% & 28.85\% \\
Insufficient hedging & 13.07\% & -- \\
Rating category definition/criterion & 23.62\% & -- \\
Relevance dispute & -- & 23.08\% \\
Temporal inconsistency & 9.55\% & 1.92\% \\
World knowledge & 25.63\% & 15.38\% \\ 
\bottomrule
\end{tabular}}
\caption{Percentage of steps per complexity category
in \dataset-Open. A step can match multiple categories. For more details see \Cref{app:data-collection}.}\label{tab:indecisive-cases}
\end{table}

The most frequent complication categories for attribution steps are world knowledge or general inference, rating category definition/criterion, and specialized knowledge and insufficient nuance/hedging tied. For logical steps, the most frequent categories are calculation or unit issue, inconsistent/unclear reference or standard, and invalid inference in a previous step.

We conjecture that improvements to the annotation instructions could help to minimize some of the categories, despite our efforts to make revisions during the pilot phases to achieve this effect. In particular, additional guidance can assist decisions on when and how to apply world knowledge, and clarify distinction between attribution labels. Recruiting annotators with appropriate domain knowledge for the main subjects of a given dataset (e.g., sports or sciences) could also reduce the number of disagreements related to both the content and the standards of comparison or equivalence.

\section{Experiments} \label{sec:baselines}

We use \dataset to evaluate existing methods for CoT verification, on step-level verification (\S\ref{subsec:step-tasks-results}) and CoT-level verification (\S\ref{subsec:full-answer-results}).

\paragraph{Verifiers.}
We use Flan-UL2 (20B), Flan-PaLM (540B), PaLM-2-L \cite{anil2023palm} and GPT-3 (\verb+text-davinci-003+) for LM baselines in few-shot prompting settings (prompt templates detailed in \Cref{app:experiments}). In addition, we use two specialized baselines: a T5-based model
with 11B parameters trained on a mixture of NLI datasets \cite{honovich-etal-2022-true-evaluating}, and FacTool---a GPT-3-based fact-checking pipeline \cite{chern2023factool}. 

For the prompting-based baselines, we use the label generated by the LM as the predicted class (which always matched to one of the task labels in our experiments). The NLI classifier receives as input a premise and hypothesis and predicts a score between 0 and 1 that indicates if the hypothesis is entailed by the premise or not. FacTool returns a binary factuality classification label, and we use it only for the attribution task.

\paragraph{Evaluation.}
As few-shot evaluation is noisy \cite{DBLP:journals/corr/abs-2105-11447,jacovi-etal-2023-comprehensive}, we average the predictions of the LM baselines over 5 different 8-shot prompts, sampled in random order from 13 demonstrations for each sub-task, which we wrote for this purpose 
(the number of demonstrations is trimmed if it exceeds a given model's context length). All sets of demonstrations were class-balanced and balanced across source datasets and labels. Prompt templates are available in \Cref{app:experiments}. For FacTool, we insert the appropriate evidence from the dataset, instead of allowing FacTool to retrieve evidence separately, for compatibility with the attribution labels. We release the prompt demonstrations alongside the dataset.

\subsection{Step-level Verification} \label{subsec:step-tasks-results}

Given the unbalanced class distribution, we report macro-F1 performance in \Cref{tab:step-tasks} (class F1 results are in \Cref{app:experiments}). 
Results for the step relevance task (classifying if a step is relevant for answering the question) are omitted, as all the models collapsed to the majority baseline, leading to an F1 of near-0.

\begin{table}[t]\centering
\setlength\belowcaptionskip{-8px}
\setlength\tabcolsep{3pt}
\scriptsize
\rowcolors{1}{}{lightgray} \ra{1.3}
\resizebox{0.99\linewidth}{!}{
\begin{tabular}{lcccc}\toprule
\textbf{Baseline} & \textbf{${\text{Attribution} \atop \text{2-class}}$} & \textbf{${\text{Attribution} \atop \text{3-class}}$} & \textbf{Logic} & \textbf{Type} \\\midrule
Flan-UL2-20B & 65.2 & 50.4 & 59.4 & 27.3 \\
Flan-PaLM-540B & 85.1 & 66.0 & 68.6 & 51.2 \\
PaLM-2-L & 85.9 & \textbf{70.7} & \textbf{77.6} & \textbf{64.1} \\
GPT-3 & 81.4 & 51.3 & 59.4 & 52.3 \\
FacTool & 71.1 & -- & -- & -- \\
t5-xxl-true & \textbf{88.4} & 55.0 & 47.3 & -- \\\midrule
Class balance & 76:24 & 70:24:6 &80:20 & 59:40:1 \\
\bottomrule
\end{tabular}}
\caption{Macro-F1 for all step-level tasks in \Cref{sec:baselines}. Performance is measured across steps.
}\label{tab:step-tasks}
\end{table}

\paragraph{Step attribution.} In this task, the model receives as input a decontextualized version of a CoT step and a Wikipedia evidence paragraph. There are two variants: \textit{2-class}---classifying whether the step is entailed or not by the evidence; \textit{3-class}---classifying between full entailment, contradiction, or not enough info. The best-performing model in 2-class was the T5-based classifier, which is much smaller than the other models although fine-tuned on large NLI data. 3-class performance follows scaling laws, which shows that detecting contradictions is a difficult problem.

\paragraph{Step logic.} This task is to classify if a given step is logically entailed by the previous steps and the question. All prompting-based baselines were biased towards a ``logically correct'' classification, achieving high F1 on the ``correct'' class ($>$85\%) and low F1 on the ``incorrect'' class (\texttildelow 33-47\%). The T5 NLI baseline shows a bias towards the ``incorrect'' class, but performs significantly worse overall. This indicates that NLI fine-tuning with simple factual claims does not generalize to out-of-domain complex logical structures.

\paragraph{Step type.} The task here is to classify whether a CoT step is an attribution step, logic step, or both, given the input question and its full answer. All models struggled with this task, achieving macro-F1 below 65\%.

\begin{table}[t]\centering
\setlength\belowcaptionskip{-8px}
\setlength\tabcolsep{3pt}
\scriptsize
\rowcolors{1}{}{lightgray}\ra{1.3}
\resizebox{0.95\linewidth}{!}{
\begin{tabular}{p{2.2cm}cc}\toprule
\textbf{Baseline} & \textbf{Single decision} & \textbf{Pipeline} \\\midrule
Flan-UL2-20B & 41.5 &54.4 \\
Flan-PaLM-540B & 39.4  & 58.1 \\
PaLM-2-L & 61.9 & \textbf{76.4} \\
GPT-3 & 35.6& 71.9 \\
\bottomrule
\end{tabular}}
\caption{Macro-F1 for the CoT correctness task (\Cref{sec:baselines}). This task involves \textit{only} specifying whether a CoT is correct or not, regardless of the exact error. The pipeline variants use the step-level decisions (relevance, type, attribution and logic tasks) to check whether an incorrect step exists in the CoT. }\label{tab:full-answer-results2}
\end{table}

\subsection{CoT-level Verification} \label{subsec:full-answer-results}
This task involves verifying whether the CoT correctly justifies the answer or not---without necessarily specifying which step contains which type of error.
Results are in \Cref{tab:full-answer-results2}. We report on two implementations of each LM baseline: (1) \textit{Pipeline} implementations are decisions by combining the LM's decision on each step separately according to the step-level task predictions in \Cref{subsec:step-tasks-results}. (2) \textit{Single-decision} implementations are by  prompting the LM to classify (in few-shot) whether a CoT is a correct answer to a question or not, given the question, CoT answer, and evidence paragraphs. The class balance for this task is 80:20 (incorrect CoTs being the majority class).

When observing macro-F1 performance, the pipeline variants significantly outperform single-decision in all cases. The gap appears to stem from a significantly higher F1 on the incorrect class (35-55\% vs. 80-87\%), while the correct class receives low F1 for both variants (30-56\%).
Overall, \textit{all baselines struggled with the high-level task of CoT verification}, showing significant room for improvement on \dataset and on CoT verification.

\section{Related Work} \label{subsec:related-work}

While CoT prompting \cite{wei2023chainofthought} has originally emerged as a simple prompting technique to increase performance of the ``final'' answer to the question, it has subsequently been interpreted as a part of the answer to be verified for correctness itself \cite{DBLP:conf/iclr/GolovnevaCPCZFC23}. As part of this narrative, various methods have been proposed to verify reasoning chains, either for LM evaluation \cite{Welleck2022NaturalProverGM}, improvement \cite{chen2023reconcile} or for training \cite{lightman2023lets}.
To our knowledge, no current work provides a fully labeled benchmark of reasoning chain correctness in information-seeking tasks. 

Recent work on verification of reasoning chains primarily focuses on methods of verification and metrics: \citet{DBLP:conf/iclr/GolovnevaCPCZFC23} propose a suite of metrics for CoT quality, including knowledge attribution and logical correctness (both formalized as instances of NLI), grammaticality, informativeness, and so on. They provide a set of examples in math QA, closed-domain QA, and NLI, annotated in various quality metrics. \citet{prasad-etal-2023-receval} propose evaluating for informativeness and logical correctness in math QA and closed-domain QA settings.
\citet{thoughtsource} collect a variety of existing gold-reference human-written CoTs from existing datasets, and provide a web-interface annotation tool for annotating CoTs for errors in logical inference, factual knowledge, verbosity and reading comprehension. \citet{lightman2023lets} provide a large dataset of math QA with step-level annotations for mathematical correctness of CoTs by one LM, primarily for training.

\section{Conclusion}

We design a methodology for human verification of reasoning chains, and employ it to annotate a dataset of CoT-format reasoning chains generated by three LMs in open-domain complex-reasoning QA tasks.
The dataset, \dataset, can be used to benchmark automatic verifiers of LM reasoning.
Our work advances the research towards LMs that can provide correct and attributable reasoning behind their decisions.
We find that CoTs generated by LMs are often not fully correct, and that automatic verifiers struggle to verify them appropriately. In particular, between attribution and logic types of correctness, \textit{CoT-generating LMs} struggle more with attribution, but in contrast, \textit{CoT verifiers} struggle more with verifying logical correctness.

\section{Limitations}

\paragraph{Evidence retrieval.} Our work focuses on evaluation of verifiers that receive evidence as input---i.e., in the task of attribution to a given source, rather than the task of fact-checking where evidence retrieval is explicitly part of the task. As such, we can only evaluate verifiers that also operate on specific given evidence, rather than fact-chekers that perform evidence retrieval themselves. Additionally, for some of the knowledge claims in the dataset labeled as ``unsupported'', it is possible that there exists evidence that will support or contradict them, which our retriever did not surface. We note, however, that the goal of the dataset is to simply collect a sufficiently wide variety of cases on which to evaluate verifiers, rather than to evaluate the CoT itself using an ``ideal'' retriever, and the labels in the dataset are well-defined for the specific evidence passages used.

\paragraph{Fermi and Wikipedia evidence.} We retrieve evidence against Wikipedia, in the interest of having clean and reliable evidence, and due to the fact that StrategyQA, MuSiQue and Sports Understanding are sufficiently addressed by Wikipedia knowledge. However, Fermi is a dataset which is explicitly designed to require knowledge that is difficult to attribute or measure, and require difficult inference for both knowledge and logic claims. As such, Fermi is by design not well-addressed by Wikipedia retrieval. We chose Fermi as one of our four datasets in order to have sufficient variety in the data for claims that are very difficult to support.

\paragraph{Reasoning chains in formats other than Chain-of-Thought.} Our dataset uses CoT as a writing style for reasoning chains, due to its popularity and sentence structure that allows convenient step separation. It is of course possible to reason about complex questions in other ways, such as using sentences that combine multiple facts together, or claims that combine knowledge and logic together, and so on. Such examples can be considered a more difficult case for reasoning verification, as they will potentially require an additional solution for extracting atomic claims from the reasoning chain. CoT-format chains, such as in our dataset, do not have this requirement.

\section*{Acknowledgments}

We thank Eran Ofek for his assistance in providing expert annotations, and to Idan Szpektor and Avi Caciularu for providing helpful feedback throughout the project. We are grateful to the annotators who provided labeling judgments and useful feedback throughout the annotation process.

\bibliography{anthology,custom}

\clearpage

\appendix

\section{\dataset: Extended Details}
\label{app:data-collection}

Here we provide additional details for the collected dataset and the collection process.

\subsection{Source Data Collection}

\Cref{tab:source-data-examples} shows examples from each of the four source datasets, alongside example answers and their correctness. These examples are shown here for illustration, and were used as part of the pilot annotation phases, so they are not examples from the final \dataset dataset.

To generate CoT answers, we constructed a separate prompt for each dataset, using examples from its training set with verified (correct) CoTs. StrategyQA, MuSiQue and Fermi provide gold-reference solutions (in non-CoT format) that we used to write the prompt CoT demonstrations. For Sports Understanding, we wrote the CoT demonstrations given the gold answer from the dataset. The CoT prompts used 6 demonstrations (question and CoT answer pairs) each.

CoT answers were split into sentences using NLTK's sentence tokenizer \cite{bird2009natural}, and each sentence was considered one reasoning step. For purposes of retrieval they were decontextualized with the decontextualization model by \citet{choi-etal-2021-decontextualization}. The evidence paragraphs were retrieved from a 2021 image of Wikipedia. The evidence paragraphs in the dataset have on average 103 words, and the longest paragraph has 582 words.

\begin{figure}[t]
\setlength{\belowcaptionskip}{-10pt}
\centering
\includegraphics[width=0.99\linewidth]{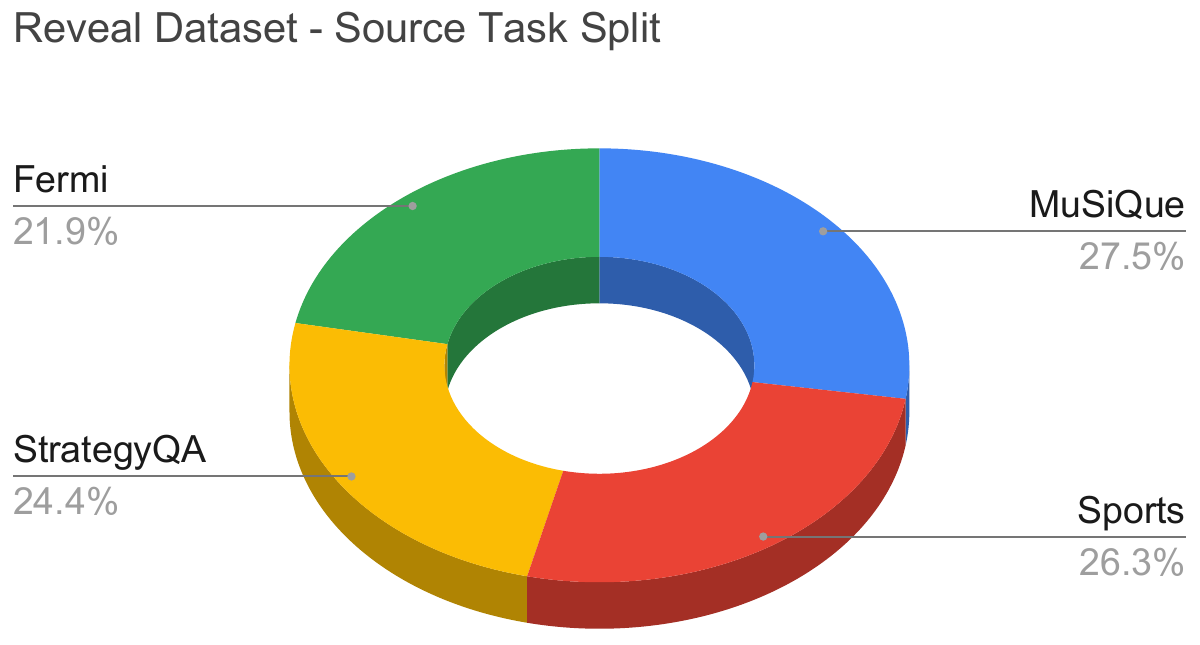} \\
\includegraphics[width=0.99\linewidth]{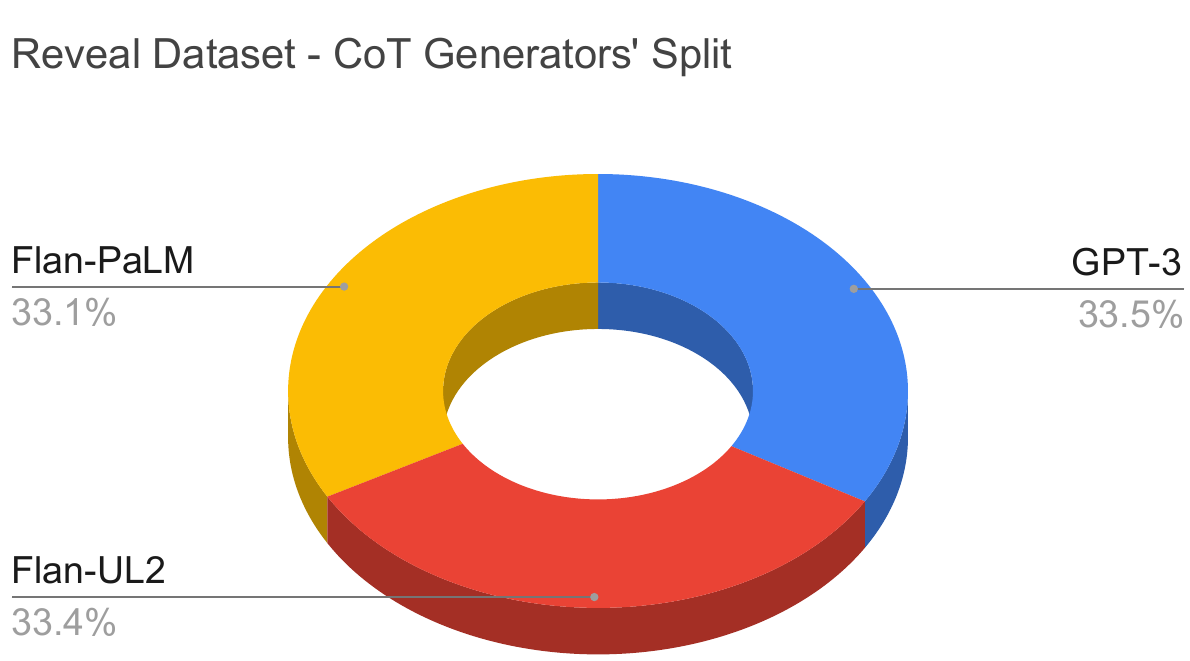}
\caption{Distributions in \dataset for the source tasks and CoT generators, which are approximately balanced.
}
\label{fig:splits}
\end{figure}

\begin{table*}[t]\centering
\scriptsize
\rowcolors{1}{}{lightgray}\ra{1.3}
\resizebox{0.99\linewidth}{!}{
\begin{tabular}{l>{\raggedright}p{3.5cm}>{\raggedright}p{5.1cm}p{2cm}}\toprule
\textbf{Source Data} & \textbf{Question} & \textbf{Answer} & \textbf{Error} \\\midrule
StrategyQA & Would a black widow have use for peaches? & A black widow is a spider. Peaches are a fruit. Thus, a black widow would have use for peaches. So the answer is yes. & Logic (step 3)  \\
StrategyQA & Will more people go in and out of Taco Bell than a Roy Rogers each year? & Taco Bell is a fast food chain with thousands of locations. Roy Rogers is a fast food chain with fewer than 500 locations. Thus, more people will go in and out of Taco Bell than Roy Rogers each year. So the answer is yes. & Correct \\
MuSiQue & Who is the other child of Chian Hsiao-chang's father? & Chian Hsiao-chang's father is Chian Ching-kuo. Chian Hsiao-chang's other child is Chian Ching-wen. So the final answer is: Chian Ching-wen. & Attribution (step 2), logic (step 3) \\
MuSiQue & Which New York City team does not host NBA games at the site where Jimmy Carter was nominated for president? & Jimmy Carter was nominated for president at Madison Square Garden in New York City. Madison Square Garden is home to the New York Knicks, but not the Brooklyn Nets. So the final answer is: Brooklyn Nets. & Correct\\
Sports &  Is the following sentence plausible? ``Jamison Crowder changed direction in the backfield.'' & Jamison Crowder is a baseball player. Changing direction in the backfield is part of American football. So the answer is no. & Attribution (step 1)  \\
Sports & Is the following sentence plausible? ``Yoan Moncada beat the buzzer.'' & Yoan Moncada is a baseball player. Beating out the buzzer is part of basketball. So the answer is no. & Correct \\
Fermi & What is the volume, cc, of blood in a healthy new-born baby? & The mass of blood in a new born baby is 25 grams. The density of blood is 0.994 g*cm**-3 or grams per cc. Thus, the cc volume of blood in a new-born baby is 25g / 0.994 g*cm**-3 = 25.15 cm**3. So the final answer is: 25.15 & Attribution (step 1) \\
Fermi & What is the total US salary? & The per capita income in the US is 63051 USD. There are around 123e+6 full time employees in the US. Thus, the total US salary is \$ 63051 * 123e+6 = 7755273000000 USD. So the final answer is: 7755273000000 & Correct \\
\bottomrule
\end{tabular}}
\caption{Examples from each of the source datasets, alongside CoT answers, and their type of error.}\label{tab:source-data-examples}
\end{table*}

\subsection{Annotation Questionnaire}

\begin{figure*}[t]
\setlength{\belowcaptionskip}{-10pt}
\centering
\includegraphics[width=0.96\linewidth]{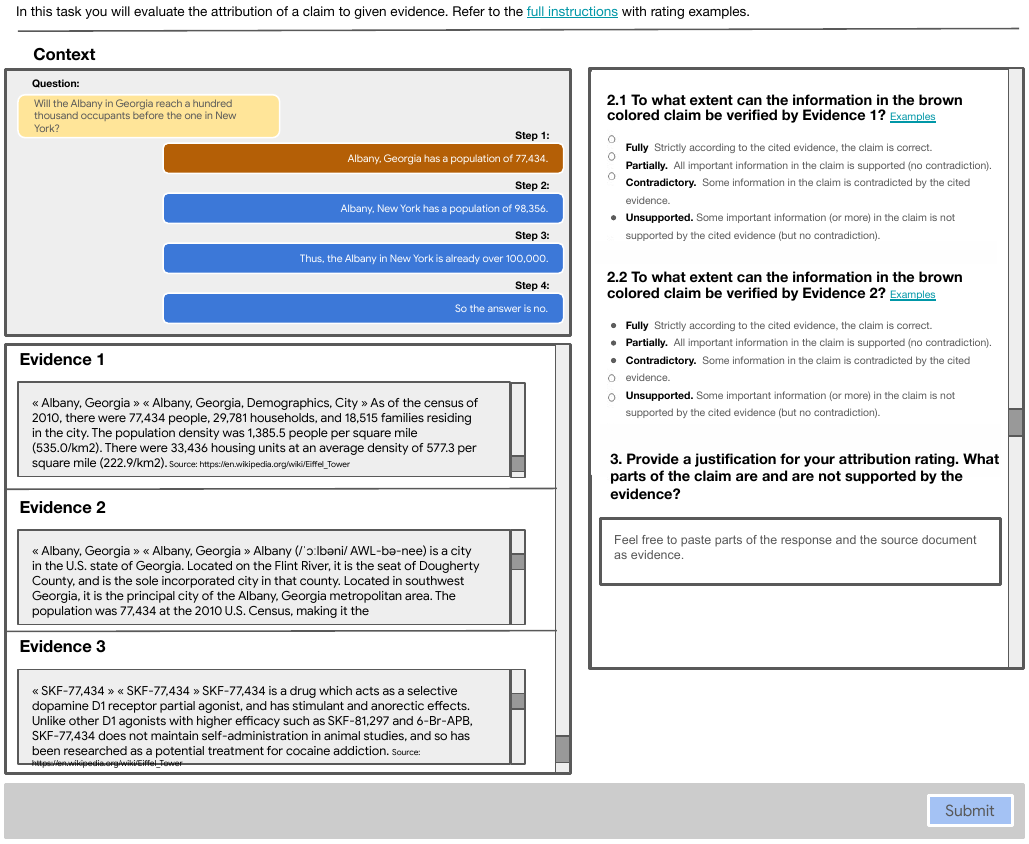}
\includegraphics[width=0.96\linewidth]{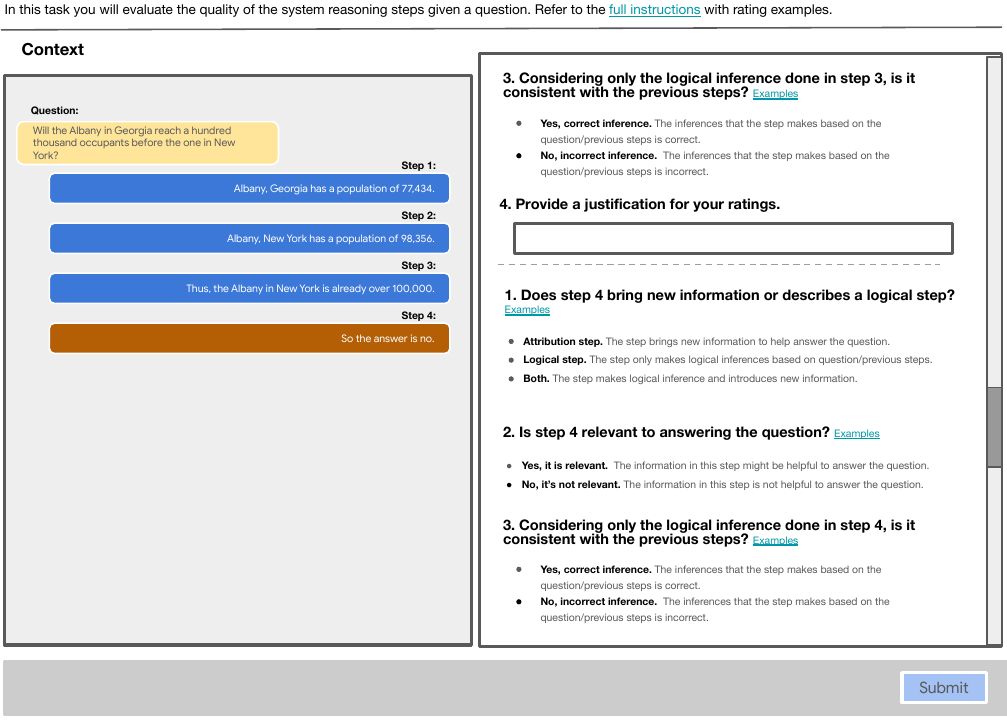}
\caption{The annotation interface for the two tasks: Attribution (above) and logic (below).
}
\label{fig:gui}
\end{figure*}

We split the annotation into two tasks, one focused on the logic annotation (including relevance, step type and logical correctness ratings), and the other focused on the attribution annotations (including relevance and step-evidence attribution). The annotation interfaces are shown in \Cref{fig:gui}.

\vspace{0.15cm}
\noindent
\textit{Task 1} is implemented and phrased as follows:
\begin{enumerate}
    \item \textbf{Relevance:} \textit{Is step $i$ relevant to answering the question?} (A) Yes, it is relevant. The information in this step might be helpful to answer the question. (B) No, it's not relevant. The information in this step is not helpful to answer the question.

\item \textbf{Type:} \textit{Does step $i$ bring new information or describes a logical step?} (A) Attribution step. The step brings new information to help answer the question. (B) Logic step. The step only makes logical inferences based on the question or previous steps. (C) Both. The step introduces new information and makes a logical inference.

\item \textbf{Logic:} \textit{Considering only the logical inference done in step $i$, is it consistent with the previous steps?} (A) Yes, correct inference. The inferences that the step makes based on the question or previous steps is correct. (B) No, incorrect inference. The inferences that the step makes based on the question or previous steps is incorrect.

\item \textbf{Justification:} \textit{Provide a justification for your ratings.}
\end{enumerate}

\textit{Task 2} is implemented and phrased as follows:

\begin{enumerate}
    \item \textbf{Relevance:} \textit{Should claim $i$ be attributed?} (A) Yes, it’s necessary to attribute the claim.  The claim has information that needs to be verified. (B) No, it’s unnecessary to attribute the claim. The claim doesn’t have any information that requires verification. 
    \item \textbf{Attribution:} \textit{To what extent can the information in claim $i$ be verified by evidence $j$?} 
(A) Fully. Strictly according to the cited evidence, the claim is correct.
(B) Partially. All important information in the claim is supported (no contradiction), but some information is unsupported.
(C) Contradictory. Some information in the claim is contradicted by the cited evidence.
(D) Unsupported. Some important information (or more) in the claim is not supported by the cited evidence (but no contradiction). 
\item \textbf{Justification:} \textit{Provide a justification for your attribution rating. What parts of the claim are and are not supported by the evidence?} Feel free to paste parts of the response and the source document as evidence.
\end{enumerate}

\subsection{Additional Statistics}

Distributions across source tasks and models are shown in \Cref{fig:splits}. To allow for a method of comparing multiple CoT answers to the same question, there are 161 questions which were answered by all three models in the dataset. Of them, 96 have all three CoT answers in \dataset-Eval, while the remainder 65 questions have one or more CoT answer in \dataset-Open.

Each step's annotation took between 300 to 600 seconds, with some answers having 5 or more steps, and with 5 annotations per step---leading to an expensive annotation process.  

\subsection{Analyses} \label{app:unsupported-analysis}

\Cref{tab:unsupported-analysis} contains the full analysis for the properties of unsupported attribution steps.

\section{Experiments and Analyses}
\label{app:experiments}

\subsection{Experiment Details}
In the few-shot evaluation settings, prompts were constructed by randomly selecting 8 demonstrations from class-balanced sets of 13 demonstrations. Prompts were trimmed to the biggest number of demonstrations which, combined with the query, were under the context length for the given LM.

The demonstrations were taken from the training sets of the source datasets, or from the examples used for the pilot annotation phase (which were discarded from the final \dataset dataset).

The dataset used to train the T5-based classification baseline is a compilation of multiple datasets: MNLI \cite{williams-etal-2018-broad}, SNLI \cite{bowman-etal-2015-large}, FEVER \cite{thorne-etal-2018-fever}, SciTail \cite{Khot_Sabharwal_Clark_2018} and VitaminC \cite{schuster-etal-2021-get}.

\subsection{Additional Results}

Per-class F1 metrics in all settings are shown in \Cref{tab:step-attribution-f1,tab:step-logic-f1,tab:full-answer-f1}.

\subsection{Prompt Templates}

We describe here the template structures we used for our few-shot verification prompts. We note that we made an effort to test multiple templates and settings, to make sure that an adequate effort was allocated to implementing baselines that are as strong as possible.

\paragraph{Attribution Task}

\begin{quote}
    Premise: [evidence paragraph]
    
    Hypothesis: [CoT step]
    
    Output: \{Entailment, Not Entailment\}
\end{quote}

\paragraph{Logic Task}

\begin{quote}
    Premise: [previous CoT steps]
    
    Hypothesis: [CoT step]
    
    Output: \{Correct, Incorrect\}
\end{quote}

\paragraph{Step Type Task}

\begin{quote}
    Question: [question]
    
    Answer: [full CoT]
    
    Step: [CoT step]
    
    Step type: \{Attribution, Logic, Both\}
\end{quote}

\paragraph{Relevance Task}

\begin{quote}
    Question: [question]
    
    Answer: [full CoT]
    
    Step: [CoT step]
    
    Is this step relevant to the answer? \{Yes, No\}
\end{quote}

\begin{table}[t]\centering
\scriptsize
\rowcolors{1}{}{lightgray} \ra{1.2}
\resizebox{0.99\linewidth}{!}{
\begin{tabular}{lrrr}\toprule
\multirow{2.5}{*}{\textbf{Baseline}} &\multicolumn{2}{c}{\textbf{F1 (Attribution Task)}} \\\cmidrule{2-3}
&Fully attributable &Not fully attributable \\\midrule
Flan-UL2-20B &51.9 &70.3  \\
Flan-PaLM-540B &78.8 &91.3 \\
PaLM-2-L & 80.0 & 91.9 \\
GPT-3 &74.5 &88.3  \\
FacTool &62.5 &79.7  \\
T5-xxl-true &83.0 &93.8  \\
\bottomrule
\end{tabular}}
\caption{Baseline results for the attribution task. The models classify each step-evidence pair. 
}\label{tab:step-attribution-f1}
\end{table}

\begin{table}[t]\centering
\scriptsize
\rowcolors{1}{}{lightgray}\ra{1.3}
\resizebox{0.99\linewidth}{!}{
\begin{tabular}{lrrr}\toprule
\multirow{2.5}{*}{\textbf{Baseline}} &\multicolumn{2}{c}{\textbf{F1 (Logic Task)}} \\\cmidrule{2-3}
 &Logically correct &Logically incorrect  \\\midrule
Flan-UL2-20B &85.1 &33.8 \\
Flan-PaLM-540B &90.2 &47.1  \\
PaLM-2-L & 90.3 & 64.9 \\
GPT-3 &86.7 &32.2  \\
T5-xxl-true  &57.6 &37.1  \\
\bottomrule
\end{tabular}}
\caption{Baseline results for the logical correctness task. The models classify the logical correctness of each step given the previous steps.}\label{tab:step-logic-f1}
\end{table}

\begin{table}[!t]\centering
\scriptsize
\rowcolors{1}{}{lightgray}\ra{1.3}
\resizebox{0.99\linewidth}{!}{
\begin{tabular}{lrrr}\toprule
\multirow{2.5}{*}{\textbf{Baseline}} &\multicolumn{2}{c}{\textbf{F1 (Full Correctness)}} \\\cmidrule{2-3}
&Correct CoT &Incorrect CoT \\\midrule
Flan-UL2-20B \textit{(pipeline)} &29.3 &79.5 \\
Flan-UL2-20B &29.3 &54.0 \\
Flan-PaLM-540B \textit{(pipeline)} &31.9 &84.3 \\
Flan-PaLM-540B &40.2 &38.6 \\
PaLM-2-L \textit{(pipeline)} & 65.2 & 87.5 \\
PaLM-2-L & 51.4 & 72.4 \\
GPT-3 \textit{(pipeline)} &56.7 &87.2  \\
GPT-3 &37.1 &34.4  \\
\bottomrule
\end{tabular}}
\caption{Baseline results for the task of classifying the full correctness of a given CoT as an answer to a question. The pipeline variants are classifiers built on the models' answers to each sub-task (relevance, type, attribution and logic), while the non-pipeline variants simply prompt the models to classify the entire CoT.}\label{tab:full-answer-f1}
\end{table}

\begin{table*}[t]\centering
\scriptsize
\rowcolors{1}{}{lightgray} \ra{1.2}
\resizebox{0.99\linewidth}{!}{
\begin{tabular}{
>{\raggedright\arraybackslash}p{1.9cm}
>{\raggedleft\arraybackslash}p{1.0cm}
>{\raggedleft\arraybackslash}p{0.8cm}
p{8cm}
}\toprule
\multirow{2.5}{2cm}{\textbf{Complexity Category}} & \multicolumn{2}{c}{\textbf{\% steps in category}} & \multirow{2.5}{*}{\textbf{Description}} \\\cmidrule{2-3}
& Attribution (N=199) & Logic (N=52) & \\\midrule
World knowledge or general inference & 25.63\% & 15.38\% & Some annotators did not apply world knowledge or general implicit inferences that other annotators take for granted (e.g., that a civil war takes place within a single country, or that an animal described as having prey can be called a ``predator''). \\ 
Acceptable nuance/hedging & 1.51\% & --  & Disagreement between annotators on strictness of hedged approximations in the step. E.g., ``There are around 7.5e+9 people in the world'' where the evidence mentions ``7.55 billion''. \\
Insufficient nuance/hedging & 13.07\% & -- & A step mentions an approximation of some quantity in the evidence but without appropriate hedging. \\
Averages and ranges & 10.05\% & -- &  A step picks a specific point or subrange from a more general range or set of ranges given in the evidence, and annotators disagree on whether the exemplification is representative of the evidence.  \\
Calculation or unit issue & -- & 42.31\% & The step consists of a calculation or unit conversion which is difficult to verify. \\ 
Specialized knowledge & 13.07\% & 7.69\% & The step requires specialized knowledge, e.g., rules of a sports game or scientific notation, which some annotations consider as world knowledge. \\
Formatting issue & 3.02\% & -- & A problem in the annotation interface (e.g., truncation) made the claims impossible to verify. \\
Inconsistent/unclear reference or standard & 12.56\% & 30.77\% & A step makes an ambiguous reference. E.g., ``Westworld was directed by Jonathan Nolan'' is paired with evidence on ``Westworld TV series \textellipsis with Nolan directing the pilot episode''---Westworld could refer to the film, TV series, etc. \\
Inference across claims in evidence & 1.51\% & -- & Verifying the step requires drawing an implicit conclusion, which only some annotators did---e.g., the step ``Birds and mammals are different classes of animals'' is paired with evidence that describes characteristics of mammals, and mentions that they ``distinguish them from reptiles and birds'', which implies that they are in different classes. \\
Invalid inference in previous step & 0.50\% & 28.85\% & Some annotators may have found it difficult to disentangle a current step's inference from any previous inferences that are invalid. \\
Rating category definition/criterion & 23.62\% & -- & Disagreements stemming from the distinction between ``partially attributable'' and ``not attributable'' (despite guidance during the annotation instructions---more discussion below). \\
Relevance dispute & -- & 23.08\%  & Borderline-relevant steps or difficult to follow steps can lead to disagreement on the relevance label. \\
Temporal inconsistency &  9.55\% & 1.92\% & The step makes a claim that relies on a different time frame from that of the evidence or other steps in the answer (e.g., present-tense in ``Maradona is a soccer player'' where the evidence mentions ``a former professional footballer''). Annotators' acceptance of temporal inconsistencies may vary (e.g., the convention of using the present tense to refer to a retired person's most salient profession). \\
\bottomrule
\end{tabular}}
\caption{An overview of the different categories we surface for the disagreements observed in \dataset-Open.}\label{tab:indecisive-cases2}
\end{table*}

\begin{table*}[t]\centering
\scriptsize
\rowcolors{1}{}{lightgray}\ra{1.3}
\resizebox{0.99\linewidth}{!}{
\begin{tabular}{l>{\raggedright}p{3.5cm}>{\raggedright}p{1cm}p{6cm}}\toprule
\textbf{Dataset} & \textbf{Step} & \textbf{Factually Correct?} & \textbf{Analysis} \\\midrule
Fermi & There are 1000 * 100 = 10000 pennies in 1 US dollar.  & No & Wrong math  \\
Fermi & Art Beat is a free, annual event in South Bend, Indiana. & Yes & The info could be found outside of Wikipedia, the retrieved evidence was irrelevant.  \\
Fermi & The latent heat of vaporization of steam is 540 cal/g. & Yes & The info could be found outside of wikipedia, the retrieved evidence was irrelevant.  \\
Fermi & The total volume of the oceans is 1.3e+21 liters. & No & The total volume of the oceans is much more than 1.3e+12 liters. The retrieved evidence was relevant but did not mention the answer explicitly (metric conversion was required).  \\
Fermi & The average American yard is 8000 square feet. & No & Could not find evidence that mentions this number, and the retrieved evidence was irrelevant.  \\
Musique & Jayantha Ketagoda is married to his wife, Jaya. & No & Could not find evidence that mentions this claim, and the retrieved evidence was relevant to the entity but did not mention the claim.  \\
Musique & Vibullia Alcia Agrippina's child was a proponent for the movement called the ''Feminist Movement''. & No & Could not find evidence that mentions this claim. \\
Musique & The Almost are signed to the record label called Almost Records. & No & The claim is wrong, and the retrieved evidence was not relevant. Could find evidence in wikipedia that refutes the claim.  \\
Musique & The most Champions League wins between 1992 and 2013 were by the Spanish La Liga. & Yes & The claim is correct, but requires complex reasoning (understanding which teams are from the Spanish La Liga). The retrieved evidence was related to La Liga but not sufficient to support the claim.  \\
Musique & Deborah Estrin is a member of the Association for Computing Machinery. & Yes & The claim is correct. There was evidence that supported the claim, but this required domain knowledge (that ``ACM''=``Association for Computing Machinery'') so the raters marked it as unsupported.  \\
Sports & Doing a maradona on the defender is part of basketball. & No & The claim is wrong, “a maradona” is a soccer move, not a basketball move.  \\
Sports & Hitting the back of the rim is a common occurrence in basketball. & Yes & The claim is correct. None of the evidence supported it directly, as it is more “common knowledge”.  \\
Sports & Windmill dunk is part of basketball, not hockey. & Yes & The claim is correct, however it is a composite claim - 1) Windmill dunk is part of basketball 2) Windmill dunk is not part of hockey. The evidence supported the first claim, not the second. \\
Sports & Brown is a basketball player. & Yes & The claim is correct, but requires further decontextualization (from ``Brown'' to ``Jaylen Brown'') as it is vague in its current form, which resulted in irrelevant evidence. \\
Sports & Being out at second is part of baseball. & Yes & The claim is correct, but requires common knowledge - that “out at second” means “second base”. The retrieved evidence did not support this claim explicitly.  \\
StrategyQA & A king size bed is 76 inches wide and 80 inches long. & Yes & The step is factually correct and a google search was able to retrieve relevant evidence. The retrieved evidence was irrelevant.  \\
StrategyQA & Pacifists do not support violence, including stoning. & Yes & The step is factually correct, and can be verified with common sense. The retrieved evidence did not support this claim. \\
StrategyQA & An existential crisis is a mental health issue.
 & Yes & The step is factually correct, and can be verified with common sense. The retrieved evidence did not support this claim explicitly.  \\
StrategyQA & Lil Jon's top ranked Billboard song was "Get Low". & No & The claim is wrong since the song is not the only top-ranked song on Billboard from the artist. The retrieved evidence was relevant and partially supported the claim (saying that the song was indeed top-ranked in the chart).  \\
StrategyQA & Cricketers do not kick field goals. & Yes & The claim is correct since a field goal is a term from other sports and kicking is generally prohibited in Cricket. The retrieved evidence was irrelevant since this requires more common-knowledge and inference.  \\
\bottomrule
\end{tabular}}
\caption{Analysis for steps where no supporting evidence was found.}\label{tab:unsupported-analysis}
\end{table*}

\paragraph{CoT Correctness task}

\begin{quote}
    Evidence $i$: [evidence paragraph]

    Question: [question]
    
    Answer: [full CoT]
    
    The answer is: \{Correct, Incorrect\}
\end{quote}

All evidences are added to the demonstration for $i$ in the number of evidences (i.e., the number of attribution steps in the CoT).

\end{document}